\documentclass[letterpaper, 10 pt, conference]{ieeeconf}  

\IEEEoverridecommandlockouts                              
\usepackage[pdftex]{graphicx}
\usepackage{tabularx}
\usepackage{multirow}
\usepackage{booktabs}

\usepackage{amsmath}
\usepackage{amssymb}
\usepackage{amsfonts}
\usepackage[ruled, vlined, linesnumbered]{algorithm2e}
\usepackage{float}

\usepackage{todonotes}

\usepackage{balance}

\usepackage{graphicx, subcaption}

\usepackage{url}

\usepackage{xcolor}

\usepackage{cite}

\usepackage{colortbl}
\definecolor{color1}{rgb}{1,0.67,0.25}  
\definecolor{color2}{rgb}{0,0.49,0.56}  
\definecolor{color3}{rgb}{0.54,0.80,0.26}  

\usepackage[belowskip=-10pt, aboveskip=2pt, font=small]{caption}
\usepackage{hyperref}

\usepackage{balance}

\title{\LARGE \bf From Model to Prototype: Design and Motor-Flap Propulsion Control of a Twin-Wing Metamorphic UAV}

 \author{Anja Bosak*, Dorian Erić*, Ana Milas, Stjepan Bogdan 
 \thanks{*Authors contributed equally. All authors are with the University of Zagreb, Faculty of Electrical Engineering  and Computing, LARICS Laboratory for Robotics and Intelligent Control Systems, Unska 3, 10000 Zagreb, Croatia; {\tt\small (anja.bosak, dorian.eric, ana.milas, stjepan.bogdan)@fer.unizg.hr}}}

\begin{document}

\maketitle

\begin{abstract}
This paper presents the design and control strategy for \textit{MetaMorpher}, a metamorphic Unmanned Aerial Vehicle (UAV), capable of both spinning-wing hover and fixed flying-wing cruise flight.
Since control of the cruise configuration is comparatively well established in the literature, this paper focuses on control of the MetaMorpher in hover mode, building on the flight dynamics model and conceptual design validated in previous work. The control algorithm is implemented using one of two phase-synchronized strategies: motor propulsion, which pulses motor thrust in synchrony with the vehicle's rotation, or flap propulsion, a novel strategy that pulses the deflection of the wing-mounted elevons. We evaluate both strategies in simulation through different flight experiments. Simulation results confirm the mathematical model and show stable reference tracking, demonstrating flap propulsion as a lightweight, decoupled alternative for hover control. Experimental testing validated the vertical-dynamics propulsion model against the physical prototype, demonstrating very good steady-state agreement across different configurations.
\end{abstract}

\IEEEpeerreviewmaketitle

\section{Introduction}
\label{sec:introduction}


Unmanned aerial vehicles (UAVs) capable of both vertical 
takeoff and hovering, as well as efficient long-range cruising, are needed for tasks such as power-line monitoring \cite{Luo2023} and bridge inspection \cite{Panigati2025}, which require confined launch sites and extended range. Conventional rotorcraft and fixed-wing aircraft each satisfy only one of these requirements. Metamorphic UAVs meet both, reconfiguring their airframe between hover- and cruise-optimized shapes. However, turning from this concept into a flying, controllable vehicle requires an actuator layout and suitable control strategy.

We address this challenge with \textit{MetaMorpher}, which builds on previously developed \textit{spincopter} \cite{Orsag2011,Orsag2013}, an inherently stable UAV whose twin-wing rotating structure generates lift by spinning two wings with a pair of propellers. MetaMorpher extends this platform into a metamorphic design that folds between the spinning-wing hover configuration and a fixed flying-wing cruise configuration. Its conceptual design and a nonlinear, segmented-aerodynamics flight dynamics model have previously been presented and validated through simulation in \cite{BosakEric2026}.

Implementing position control in the hover configuration, however, poses a challenge that conventional Vertical Take-Off and Landing (VTOL) controllers do not face. In this case, the aerodynamic surfaces and propulsion units rotate with the airframe, a fixed-axis control input cannot produce a constant horizontal force. Instead, the control action should be synchronized with the vehicle's rotational phase and applied only within selected angular sectors of each revolution. The spincopter implemented this principle by pulsing the imbalance between its two motor thrusts \cite{Orsag2011,Orsag2013}, an approach that couples horizontal and vertical control since both motors also determine the vehicle's overall lift \cite{Orsag2011}. 
Since MetaMorpher's cruise mode already requires a pair of wing-mounted elevons for attitude control, a well-established problem (\cite{Hu2024}), that we do not address in detail here, motivated us to use it as an alternative channel for the novel control principle.

Therefore, this paper presents the design and control algorithm for the MetaMorpher. The resulting vehicle has a 0.7 m wingspan and an estimated ready-to-fly mass of approximately 454 g, with two independently driven wing-mounted propellers and two independently actuated trailing-edge elevons. For position control in hover mode, we implement and compare two phase-synchronized strategies: motor propulsion, which pulses the motor thrust imbalance following the spincopter approach in \cite{Orsag2011}, and flap propulsion, a novel strategy that instead pulses the elevon deflection in synchrony with the vehicle's rotation while motor speed is held constant. We evaluate both strategies using a nonlinear Simulink model, demonstrating stable waypoint tracking in simulation.
\begin{figure}[t]
	\centering
 \includegraphics[width=1.0\columnwidth]{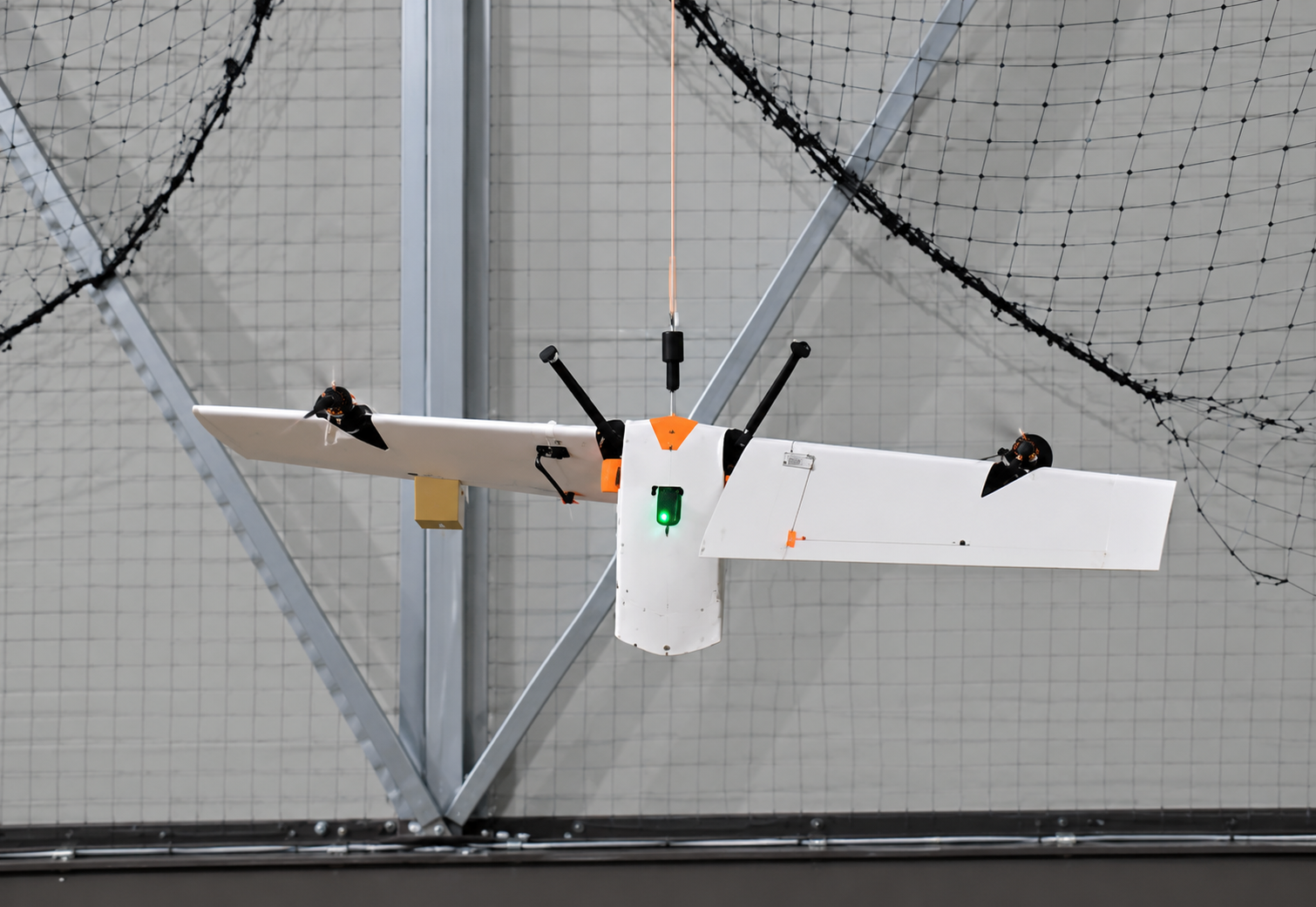}
	\caption{UAV in hover mode configuration during real-world experiments.}
	\label{fig:metamorpher_real_world}
\end{figure}
The main contributions of this work are:
\begin{enumerate}
    \item Physical design of a MetaMorpher prototype that implements the airframe and actuator layout derived from the mathematical model in \cite{BosakEric2026}.
    \item Implementation and comparison of two phase-synchronized control strategies for position control in hover mode: motor propulsion and flap propulsion, a novel scheme that repurposes MetaMorpher's cruise-mode elevons for hover-mode actuation validated through simulation tests.
    \item Parameter identification of the throttle-to-spin-rate response in simulation and on the physical prototype.
    \item An open-source Simulink code of the control strategy to support reproducibility.
\end{enumerate}

This paper is divided into six sections. Section \ref{sec:related} reviews related work on different designs and control algorithms of MetaMorpher-like vehicles. Section \ref{sec:design} describes the MetaMorpher prototype. Section \ref{sec:control} presents the motor- and flap-propulsion control strategies. Sections \ref{sec:simulation} and \ref{sec:experiment} report the simulation and experimental setup and flight-test results, respectively, to validate the design of the vehicle. Finally, Section \ref{sec:conclusion} presents conclusions and directions for future work.

\section{Related work}
\label{sec:related}

Conventional UAV airframes typically use a single configuration, trading hover capability for cruise efficiency. Metamorphic designs avoid this trade-off through approaches ranging from bio-inspired wing folding \cite{Win2021} to dedicated tilt-rotor \cite{Anuar2025} and tail-sitter \cite{Sufiyan2021} layouts. However, most of this work stops at mechanical concepts and theoretical analysis. Accurate simulation models and physical prototypes with control strategies validated through flight tests remain rare.

Few review papers address the design of morphing aircraft alongside their flight control. For instance, Chu et al. \cite{Chu2022} reviewed the broader field of morphing aircraft, covering configuration design methods, the nonlinear dynamics and aerodynamic interference introduced by wing morphing, and the flight control strategies used to address them. Authors in \cite{Parancheerivilakkathil2024} focused more specifically on control, surveying shape, flight-dynamics, and aeroelastic control strategies for a wide range of fixed-wing morphing concepts. Harvey et al. \cite{Harvey2022} took a different angle, reviewing avian-inspired morphing in which wing sweep, dihedral, twist, and camber replace conventional control surfaces to provide pitch, roll, and yaw control, including on tailless airframes; camber-morphing trailing edges, similar to elevons, provide pitch control when actuated symmetrically and roll control when actuated asymmetrically. Across these reviews, the scope remains fixed-wing or gliding flight, while none address vehicles like MetaMorpher that transition between rotary-wing hover and fixed-wing cruise.

Several hardware platforms have demonstrated metamorphic hover-to-cruise flight. Low et al. \cite{Low2017, Low2018} developed the Transformable HOvering Rotorcraft (THOR), which inspired MetaMorpher’s wing sizing and models the vehicle as a rigid body in each flight mode. Sufiyan et al. \cite{Sufiyan2021, Sufiyan2022} introduced a multimodal vehicle, jointly optimizing design and control parameters through a collaborative co-evolutionary approach. State-of-the-art control papers have also focused on specific aspects of the transition problem. Yuksek et al. \cite{Yuksek2016} developed a six-degree-of-freedom nonlinear model for transition control of a tilt-rotor UAV and validated it through both simulation and a physical flight demonstration on their TURAC prototype. Zhao et al. \cite{Zhao2021} proposed Kriging-based control allocation for a flying-wing UAV with a morphing trailing edge, validated in simulation, addressing control-surface actuation without full airframe metamorphosis or a hover mode. 

Both of MetaMorpher's control strategies, detailed in Section \ref{sec:control}, rely on synchronizing the control input with the vehicle's rotational phase rather than producing a fixed control force in the body frame. This principle is closely related to cyclic pitch control on a conventional helicopter rotor, where blade pitch is varied once per revolution to redirect the rotor's thrust in a commanded inertial direction \cite{Leishman2006}.
On the spincopter, this principle was implemented entirely through differential motor thrust \cite{Orsag2011,Orsag2013}. Phase-synchronized flap deflection has also been investigated on single-wing, samara-inspired monocopters, where a trailing-edge flap is deflected within selected angular sectors of each revolution to produce horizontal control forces \cite{FregeneBolden2010}. 
To the best of our knowledge, no existing metamorphic UAV has achieved horizontal control through elevon deflection on a twin-wing spinning platform, nor directly compared it against motor pulsation on the same model.

This work bridges these gaps by implementing MetaMorpher’s previously validated model as a flight-ready prototype and evaluating two phase-synchronized control strategies for its hover mode.  Finally, this work lays the foundation for combining motor and flap propulsion into a single control architecture and for extending phase-synchronized control across the transition between hover and cruise flight.
\section{MetaMorpher Design}
\label{sec:design}

\begin{figure}[t]
	\centering
 \includegraphics[width=1.0\columnwidth]{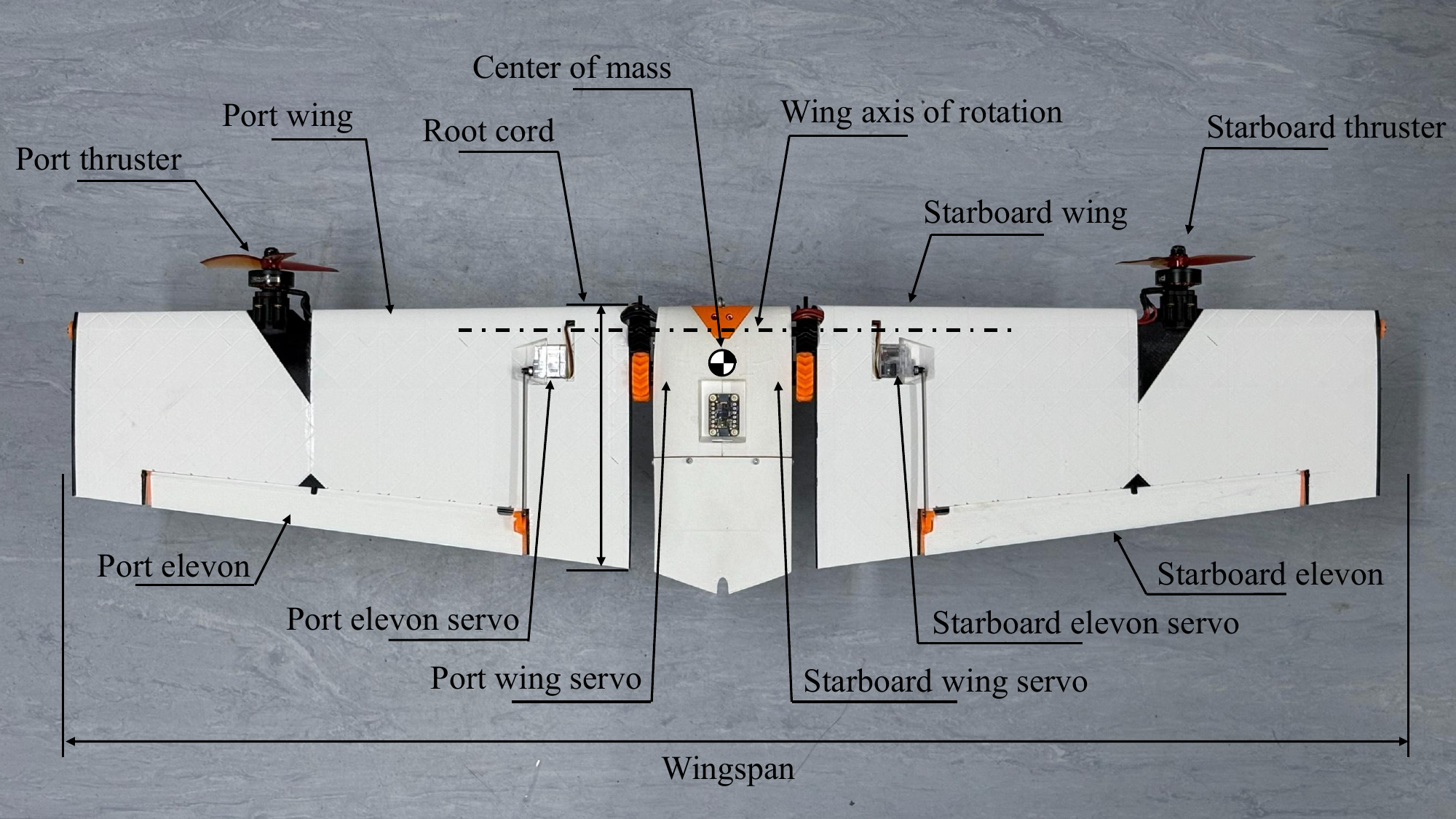}
	\caption{UAV in cruise mode configuration with marked-down actuator and control surface positions.}
	\label{fig:metamorpher_design}
\end{figure}

\subsection{Design Overview}

The UAV is designed to operate in both spinning-wing hover and fixed-wing cruise configurations. The transition mechanism follows a concept similar to \cite{Low2018}, where the two wings are independently rotated by servo actuators about spanwise pivot axes. Each wing is connected to the fuselage through a one-degree-of-freedom (1-DoF) revolute joint. The pivot shafts also serve as the primary wing spars, reducing the number of structural components and the overall mechanism complexity.

The final vehicle layout, shown in Fig.\ref{fig:metamorpher_design}, includes two independently driven wing-mounted propellers  and two independently actuated trailing-edge control surfaces, referred to as flaps in the spinning-wing hover configuration and as elevons in the fixed-wing cruise configuration, reflecting their differing aerodynamic role in each flight mode. This arrangement provides separate port and starboard propulsion and aerodynamic control inputs, which are used in this work to generate asymmetric forces and moments. Further details of the mechanical layout, electronic components, and wing geometry are given in the following subsections.

\subsection{Actuator Layout}
\label{subsec:actuator_layout}

The wing pivot shafts pass through the central fuselage and are positioned as far forward as permitted by the wing geometry and structural requirements. This leaves sufficient volume for the battery, flight controller, and other onboard electronics while allowing the center of mass to be placed within the required range. The pivot shafts also act as the primary wing spars, such that the wing rotation mechanism does not require an additional load-bearing element. The rotation of the wing is coupled to the actuator through a pair of gears, giving the wings a \(\pm 90^{\circ}\) range of motion.

The propulsion units are positioned near the outer wing sections to increase the moment arm of the thrust force about the vehicle rotation axis. This increases the available spin-up moment in the hover configuration and the control moment available from differential thrust. However, the local tangential velocity at the propeller distance from the center of the UAV, $r_m$, increases with radial distance from the rotation axis $\omega_x$, \(V_{\mathrm{tan}} = r_m \omega_x\). The motor position was therefore selected as a compromise between thrust moment and propeller performance under the increased local inflow velocity. The final actuator placement is shown in Fig.~\ref{fig:metamorpher_design}.

The elevon servo actuators are installed within recessed cavities on the upper wing surface to minimize external protrusions and the associated aerodynamic drag. A mechanical linkage transfers the servo motion to the trailing-edge elevons. The geometry of the linkage provides an elevon deflection range of \(\pm 35^{\circ}\) (with respect to the wing), which is used for both conventional control inputs and the asymmetric periodic actuation considered in this work.

\subsection{Wing Design}
The wing geometry was selected following the design procedure from \cite{BosakEric2026}. The UAV has a wingspan of \(0.7~\mathrm{m}\), a reference wing area of approximately \(0.09~\mathrm{m^2}\), a root chord of \(160~\mathrm{mm}\), and a taper ratio of \(0.688\). For the estimated ready-to-fly mass of \(0.45~\mathrm{kg}\), this corresponds to a target wing loading of \(5~\mathrm{kg/m^2}\).

The cruise configuration uses the reflexed Eppler E387 airfoil, while a symmetric NACA 0010 profile is used in the hover configuration. The symmetric hover profile avoids an aerodynamic bias when one wing is
inverted during the transformation. The two profiles were selected with similar outer geometry to support the wing-profile morphing mechanism.
Each wing contains an independently actuated trailing-edge elevon with a span of \(225~\mathrm{mm}\) and a mean chord of \(23~\mathrm{mm}\). The elevons are positioned with their inboard edges aligned with the corresponding servo arms, simplifying the linkage geometry and reducing the required linkage length. Their outboard extent overlaps the propeller slipstream, increasing the local dynamic pressure over part of the elevon and maintaining aerodynamic control at low forward speeds \cite{ikami2021low}.

\subsection{Hardware Components}

\label{subsec:electronics}

The onboard electronics consist of a custom Flight Controller (FC) based on the RP2040 microprocessor, two Electronic Speed Controllers (ESCs), a radio receiver, and an IMU module. The flight controller gathers inertial measurements from the IMU and generates independent commands for the port and starboard propulsion units, wing actuators, and elevon servos. The two propulsion units are driven by ESCs running AM32 firmware using the DSHOT protocol for communication with the FC, selected to support the current requirements of the \(2207\) BLDC motors and to allow rapid motor-speed modulation. The vehicle is powered by a \(3\)S \(1300~\mathrm{mAh}\) LiPo battery. The component mass breakdown is given in Table~\ref{tab:component_mass}.

\begin{table}[t]
\centering
\caption{Mass breakdown of MetaMorpher components.}
\label{tab:component_mass}
\footnotesize
\setlength{\tabcolsep}{4pt}
\begin{tabular}{lcc}
\toprule
\textbf{Component} & \textbf{Qty.} & \textbf{Unit mass [g]} \\
\midrule
Flight Controller             & 1 & 10  \\
IMU                           & 1 & 4  \\
Electronic Speed Controller   & 2 & 12  \\
BLDC motor                   & 2 & 33 \\
Propeller                    & 2 & 3  \\
Battery (3S, 1300 mAh)       & 1 & 108 \\
Receiver                     & 1 & 3  \\
Carbon-fibre spar            & 1 & 17 \\
Fuselage                     & 1 & 38 \\
Wing                          & 2 & 65 \\
Wing pivot actuator          & 2 & 12 \\
Elevon                       & 2 & 7  \\
Elevon actuator              & 2 & 5  \\
\midrule
\textbf{Total estimated mass} & & \textbf{454} \\
\bottomrule
\end{tabular}
\end{table}
\section{Flight Control}
\label{sec:control}

\begin{figure}[t]
	\centering
 \includegraphics[width=1.0\columnwidth]{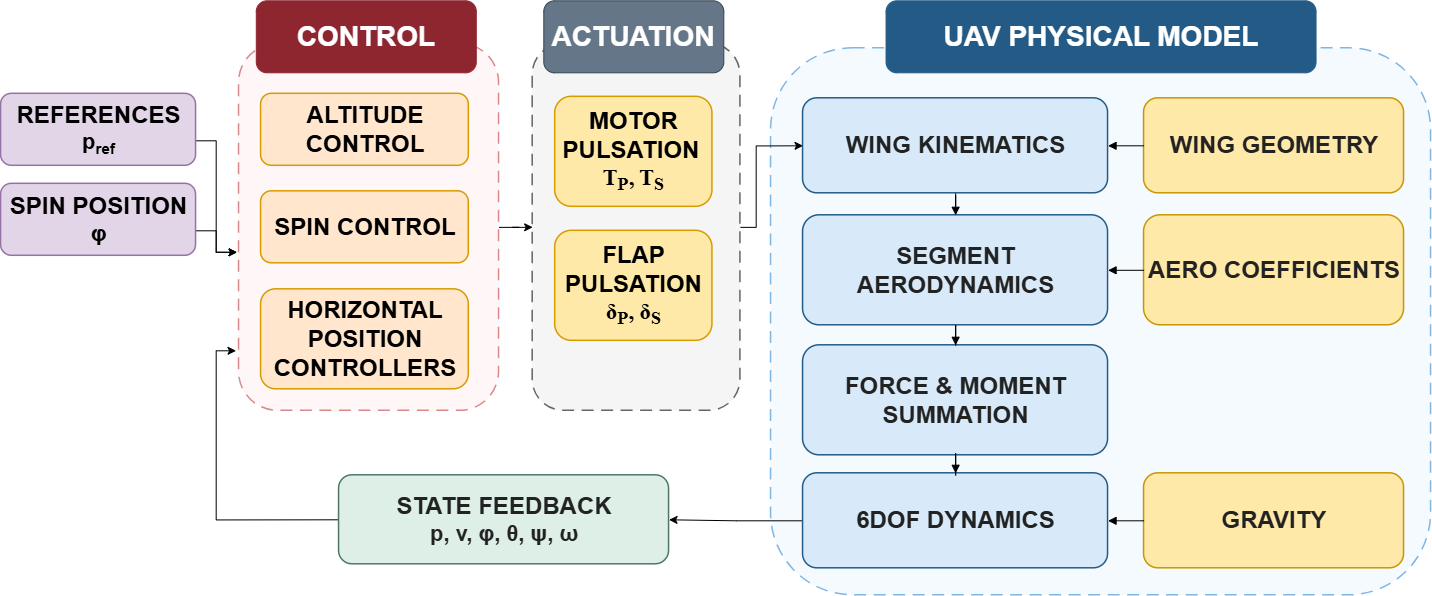}
	\caption{Overall schematic diagram of the control, actuation, and UAV model modules. The control module includes altitude, spin, and horizontal position controllers, whose outputs are propagated to motor and flap pulsation. The vehicle model sends the actuator commands through wing kinematics, segment aerodynamics, and force and moment summation to obtain the 6-DOF dynamics, parameterized by wing geometry, aerodynamic coefficients, and gravity \cite{BosakEric2026}. State feedback closes the loop back to the control module.}
	\label{fig:diagram}
\end{figure}

The flight control system was developed for the spinning flight configuration of the vehicle, in which the aerodynamic surfaces and wing-mounted propulsion system rotate together with the airframe. In this configuration, conventional fixed-axis control inputs cannot be applied directly to generate a constant force in the inertial horizontal plane. Instead, the control action must be synchronized
with the rotational phase of the vehicle.
Two control approaches were considered. The first uses differential modulation of the tip-mounted motors, while the second uses periodic deflection of the aerodynamic flaps. In both cases, horizontal control is achieved by applying the control input only during selected portions of each rotation.

\subsection{Motor Pulsation}

The motor-based control system uses the two wing-mounted propellers both to support the vehicle vertically and to generate horizontal control forces. The motor commands therefore consist of a nominal thrust component together with additional contributions from the altitude, rotational-speed, and horizontal-position controllers.

Altitude control is achieved by simultaneously adjusting the thrust of
both motors around the nominal operating point. The altitude error,
determined from the difference between the reference and measured
altitude, is processed by the altitude controller to generate a thrust
correction. When additional vertical force is required, the thrust of
both motors is increased, while a reduction in thrust is commanded when the vehicle needs to descend. Applying the altitude correction equally to both motors changes the total thrust while maintaining the required balance between the two propulsion units.

The rotational motion about the longitudinal body axis is controlled
using the measured roll rate $\omega_x$. The difference between the
reference and measured roll rate is processed by a PI controller. The
resulting control action modifies the motor commands so that the desired spinning condition can be maintained.

As already mentioned, the horizontal motion is generated using a phase-synchronized motor
pulsation mechanism. The horizontal position error is first converted
into a desired direction of motion in the horizontal plane of
$\mathcal{F}_W$. This direction is represented by a phase angle,
$\lambda$, which determines the angular location within each vehicle
rotation at which the motor correction should be applied.

The instantaneous rotational phase $\gamma$ of the vehicle is compared
with $\lambda$. A threshold-based switching law activates the horizontal motor correction only within selected angular sectors of the rotation:

\begin{equation}
\Delta T_{xy} =
\begin{cases}
+A_T, & \cos(\gamma-\lambda) \geq \eta,\\
-A_T, & \cos(\gamma-\lambda) \leq -\eta,\\
0, & \text{otherwise},
\end{cases}
\label{eq:motor_switching}
\end{equation}

where $A_T$ denotes the motor pulsation amplitude and $\eta$ is the
switching threshold. Opposite corrections are then applied to the port
and starboard motors as:

\begin{equation}
\begin{split}
T_P &= T_{\mathrm{base}} + \Delta T_{xy}, \\
T_S &= T_{\mathrm{base}} - \Delta T_{xy},
\end{split}
\label{eq:motor_mixing}
\end{equation}

where $T_{\mathrm{base}}$ includes the nominal thrust and the
contributions required for altitude and rotational-speed regulation,
while $\Delta T_{xy}$ represents the phase-dependent horizontal
control input.

Since the vehicle continuously rotates, the direction associated
with the differential motor action also rotates in the inertial frame.
Synchronizing the thrust modulation with the rotational phase causes
the individual periodic contributions to produce a non-zero average
force in the commanded horizontal direction. Changing the phase
$\lambda$ changes the direction of this average force, while changing
the pulsation amplitude modifies its magnitude.

\subsection{Flap Pulsation}

The flap controller applies the same phase-synchronized control
principle as the motor-based approach, while using periodic aerodynamic flap deflections to generate the required horizontal control action.

Let $\mathbf{p}_{xy}$, $\mathbf{p}_{xy,\mathrm{ref}} \in \mathbb{R}^2$  denote the vehicle's measured and reference horizontal position in $\mathcal{F}_W$, and $\mathbf{v}_{xy} \in \mathbb{R}^2$ its measured horizontal velocity. The position error:

\begin{equation}
\mathbf{e}_p = \mathbf{p}_{xy,\mathrm{ref}} - \mathbf{p}_{xy},
\label{eq:flap_pos_error}
\end{equation}

is converted into a limited velocity reference:

\begin{equation}
\mathbf{v}_{xy,\mathrm{ref}} =
\mathrm{sat}\!\left(K_p\,\mathbf{e}_p,\;v_{\max}\right),
\label{eq:flap_vel_ref}
\end{equation}
where $K_p$ is a proportional gain and
$\mathrm{sat}(\cdot,v_{\max})$ limits the vector magnitude to
$v_{\max}$ while preserving its direction. The resulting velocity
error:
\begin{equation}
\mathbf{e}_v =
\mathbf{v}_{xy,\mathrm{ref}}-\mathbf{v}_{xy},
\label{eq:flap_vel_error}
\end{equation}
determines the commanded flap-deflection amplitude and phase angle:
\begin{equation}
\begin{aligned}
A_f &= \mathrm{sat}\!\left(K_v\lVert\mathbf{e}_v\rVert,\,
A_{f\max}\right),
\label{eq:flap_amplitude}\\
\lambda &= \mathrm{mod}\!\left(
\lambda_X-\mathrm{atan2}(e_{vy},e_{vx}),\,
2\pi
\right),
\end{aligned}
\end{equation}
where $K_v$ is a proportional gain, $A_{f\max}$ is the maximum
commanded flap-deflection amplitude, bounded by the elevon's
mechanical deflection range (Section~\ref{sec:design} ), and
$\lambda_X$ is the calibrated actuation phase corresponding to the
positive $x$ direction.


The rotational phase of the vehicle is compared with the commanded
phase. The flap is activated only within predefined angular regions
of each revolution. The corresponding phase-synchronized flap command
is defined as:

\begin{equation}
\delta_f =
\begin{cases}
+A_f, & \cos(\gamma-\lambda) \geq \eta,\\
-A_f, & \cos(\gamma-\lambda) \leq -\eta,\\
0, & \text{otherwise},
\end{cases}
\label{eq:flap_switching}
\end{equation}

where $\eta$ is the switching threshold. The switching law therefore produces positive flap deflection in one angular sector, negative deflection in the opposite sector, and zero deflection during the remaining portion of the rotation.

Since the orientation of the aerodynamic control action changes
continuously as the vehicle rotates, this phase synchronization allows
the flap actuation to generate a control effect in the desired
horizontal direction. The phase command determines the direction of
the control action, while the flap-deflection amplitude determines its
magnitude.

\subsection{Combined Control Strategy}

The motor and flap controllers are based on the same underlying
concept, the continuous rotation of the vehicle is used as part of the control mechanism. Rather than attempting to produce a fixed body-frame control force, the control input is synchronized with the rotational phase so that the resulting force acts in a desired inertial-frame direction.


In this work, the motor and flap pulsation mechanisms are developed and evaluated independently, characterizing each channel’s individual contribution to horizontal control. Our results show that flap pulsation achieves a faster response than motor pulsation (Fig. \ref{fig:flap_position}, Fig. \ref{fig:motor_position}). This suggests that the two mechanisms are not simply interchangeable but potentially complementary. Flap pulsation can be used for fast, fine-grained position corrections, and motor pulsation as an additional control for larger position changes. Combining both actuators into a single architecture that exploits this complementarity is proposed as future work. In such a system, the motors would continue to provide the primary thrust, altitude regulation, and rotational-speed control, while horizontal control would be distributed between motor and flap pulsation according to the magnitude and urgency of the required correction.
\section{Simulation-based evaluation}
\label{sec:simulation}

The proposed control approaches were evaluated using the nonlinear MetaMorpher simulation model developed in Simulink, available at \footnote{\href{https://anonymous.4open.science/r/metamorph_uav_matlab-F10C/README.md}{MetaMorpher\_control}}. The analysis was performed in the spinning flight configuration and focused on the vertical dynamics of the vehicle and its horizontal position control capability. First, the vertical
response was identified for different wing configurations. The motor and flap pulsation controllers were then evaluated separately to determine their ability to generate controlled horizontal motion while maintaining the commanded altitude.

\subsection{Vertical Dynamics Identification}

Prior to evaluating the horizontal control strategies, the vertical dynamics of the vehicle were characterized around a steady operating condition (hovering). A small step change in motor  thrust was applied and the resulting change in vertical velocity was recorded. The response was then used to identify a second order dynamic model.

\begin{figure}[!t]
    \centering

    \includegraphics[width=0.9\columnwidth]{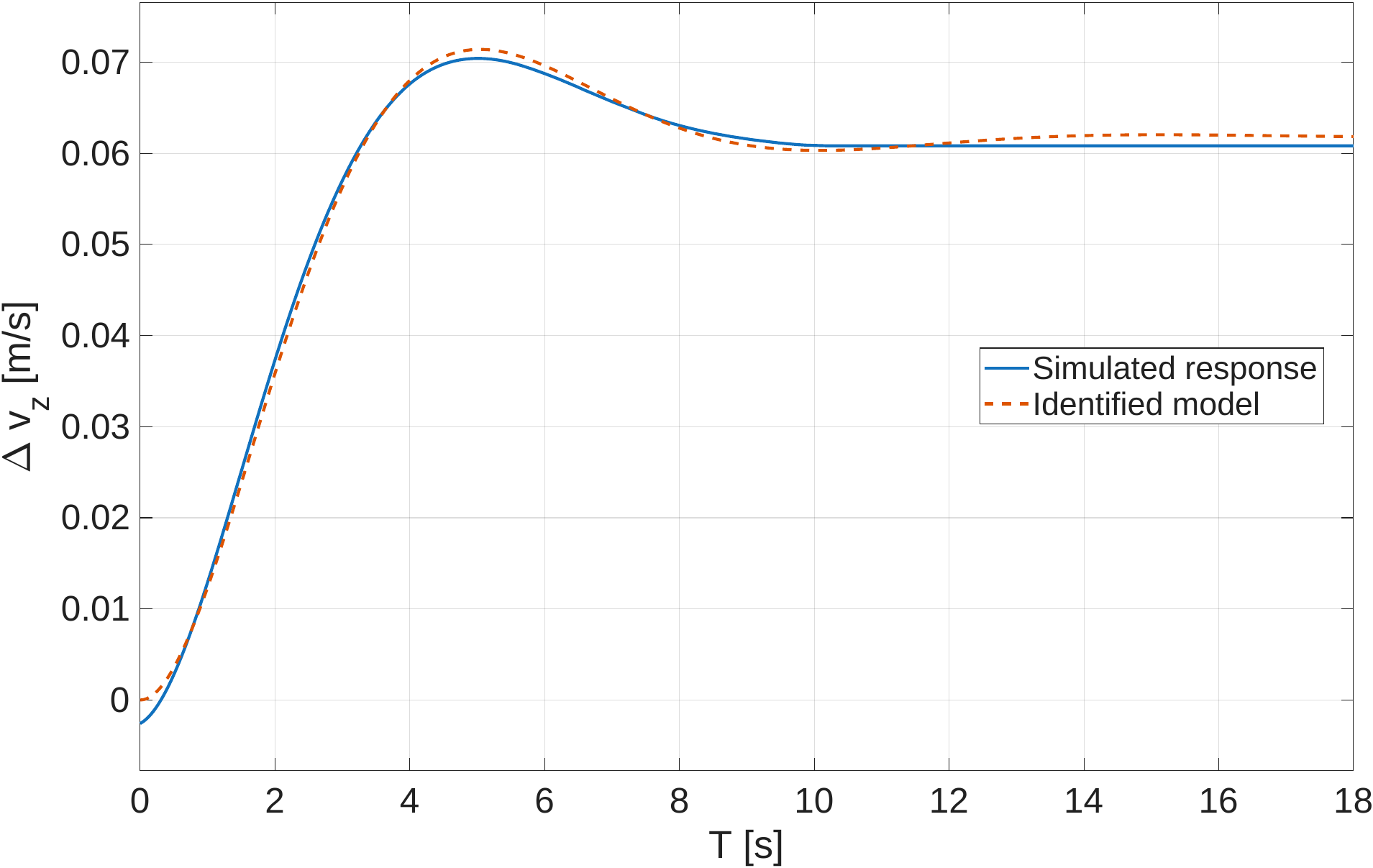}

    \vspace{0.8em}

    \includegraphics[width=0.9\columnwidth]{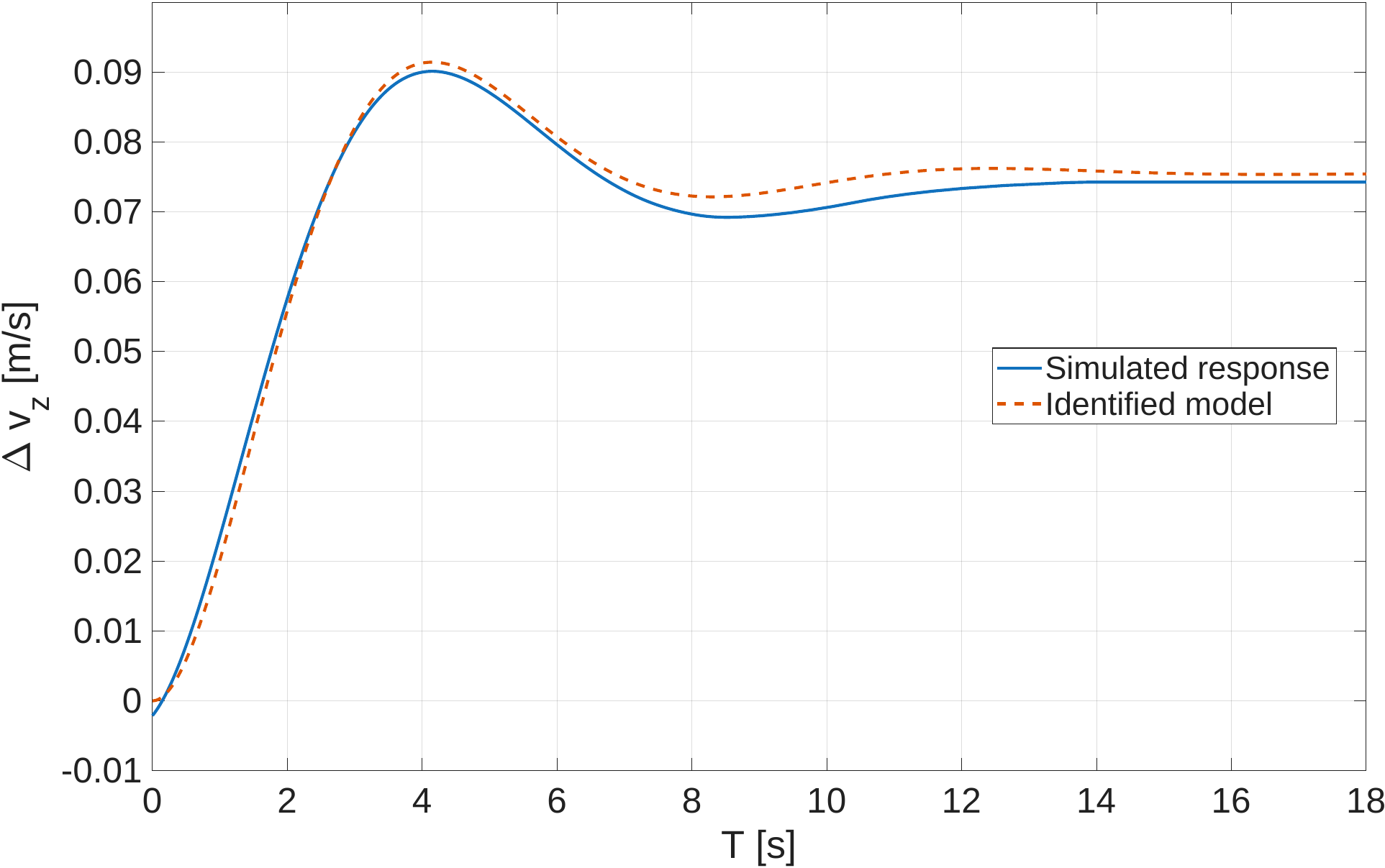}

    \caption{Comparison of the simulated vertical velocity response
    and the identified second-order models for the $30^\circ$ (top)
    and $40^\circ$ (bottom) wing configurations.}
    \label{fig:vertical_id}
\end{figure}

Fig.~\ref{fig:vertical_id} compares the simulated vertical velocity
responses with the identified models for the $30^\circ$ and $40^\circ$
wing configurations. In both cases, the vertical speed response exhibits underdamped
behavior, characterized by an initial increase in vertical velocity,
followed by an overshoot and convergence toward a steady-state value.

The similarity between the simulated and identified responses indicates
that the dominant vertical dynamics around the considered operating
points can be represented by a low-order linear model. The identified dynamics
characterize the vehicle response around the spinning-flight operating
condition and provide a basis for the development of the altitude
control loop.

\subsection{Motor Pulsation Control}

The motor pulsation controller was first evaluated using a fixed
horizontal position reference. The commanded changes in position were defined as
$p_{x,\mathrm{ref}}=0$~m and $p_{y,\mathrm{ref}}=1$~m, while the
altitude reference change was set to $p_{z,\mathrm{ref}}=1$~m.

\begin{figure}[t]
    \centering
    \includegraphics[width=0.85\columnwidth]
    {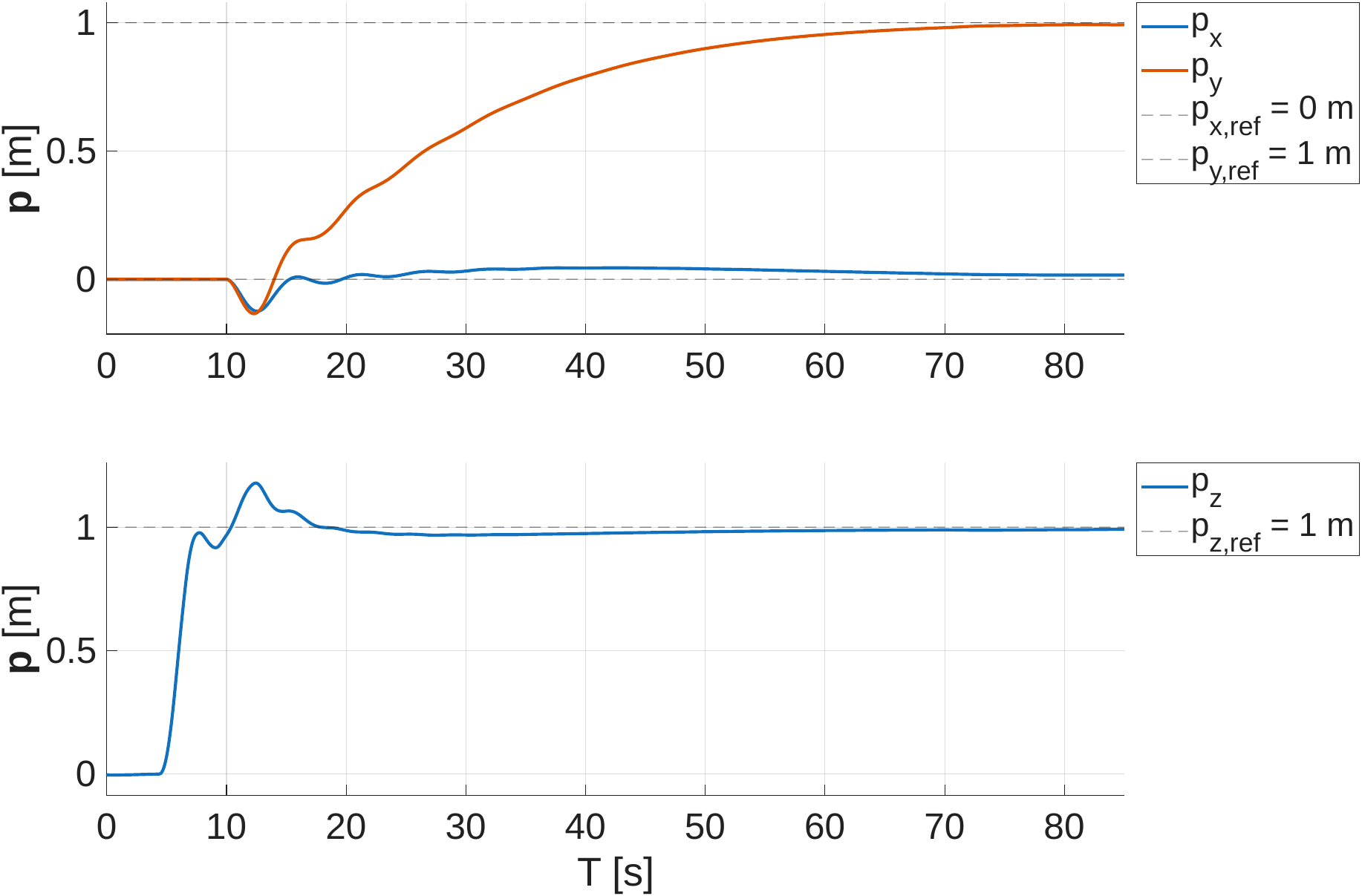}
    \caption{Position response using the phase-synchronized motor
    pulsation controller for $p_{x,\mathrm{ref}}=0$~m,
    $p_{y,\mathrm{ref}}=1$~m, and
    $p_{z,\mathrm{ref}}=1$~m.}
    \label{fig:motor_position}
    \vspace{2em}
\end{figure}

Fig.~\ref{fig:motor_position} shows the resulting position response.
Following the initial vertical transient, the altitude converges toward
the commanded value and remains close to the reference. In the
horizontal plane, the controller produces motion primarily along the
commanded $y$ direction, with $p_y$ approaching the \(1~\mathrm{m}\) reference.
At the same time, the displacement along the non-commanded $x$
direction remains comparatively small. This response demonstrates that
the phase-synchronized differential motor thrust can generate a directed
horizontal control action while altitude is kept unchanged.

To further evaluate the controller, a sequence of horizontal waypoints
was introduced together with a varying altitude reference. The
horizontal path was defined by three waypoints, $P_1$, $P_2$,
and $P_3$, requiring the controller to change the commanded direction
of motion during the maneuver. The altitude reference was simultaneously
changed from $1$~m to $1.2$~m and subsequently returned to $1$~m.

\begin{figure}[t]
    \centering
    \includegraphics[width=\columnwidth]
    {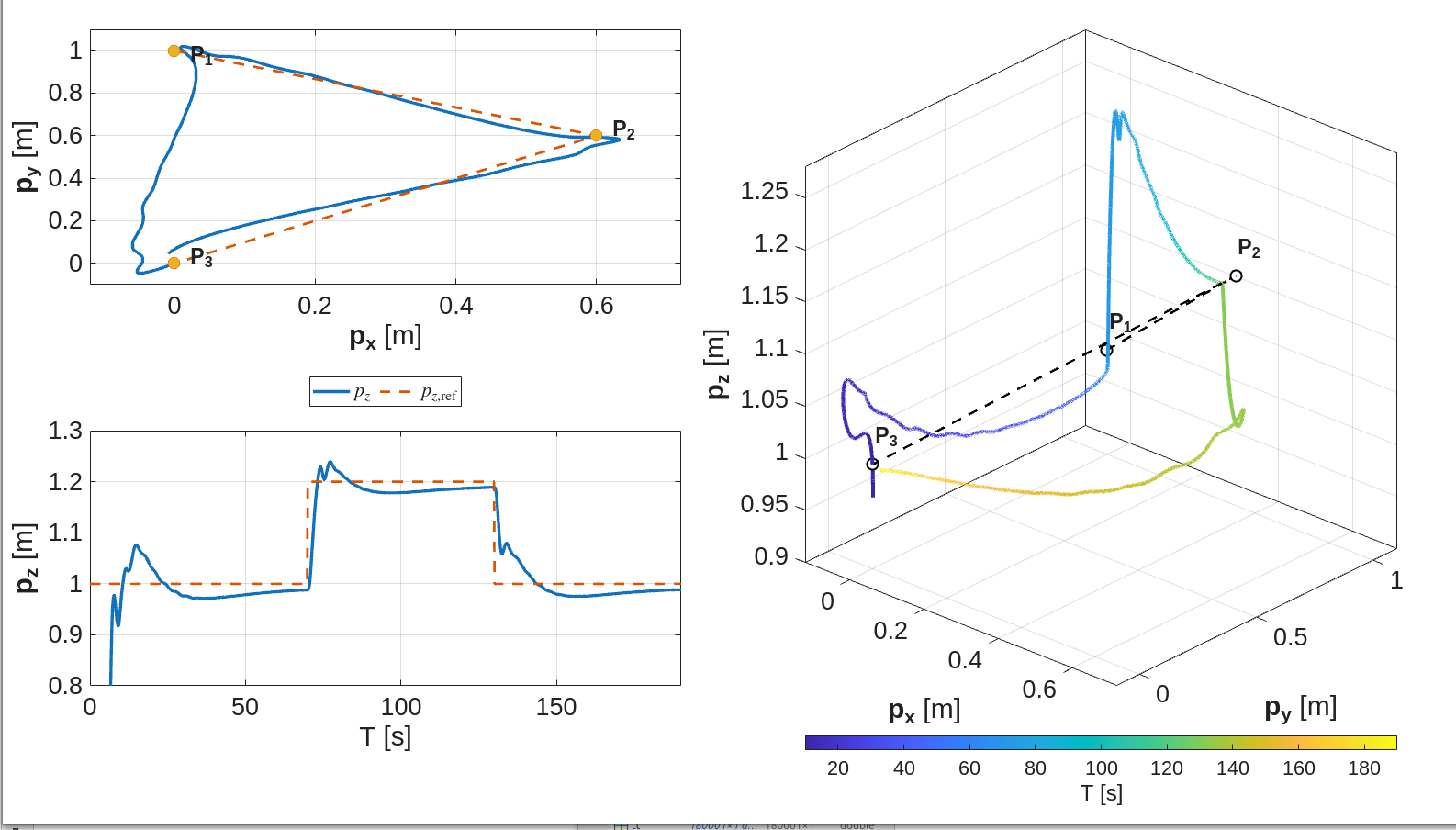}
    \caption{Waypoint tracking using the phase-synchronized motor
    pulsation controller. The horizontal trajectory is defined by
    waypoints $P_1$, $P_2$, and $P_3$, while the altitude reference
    is varied during the maneuver.}
    \label{fig:motor_trajectory}
    \vspace{2em}
\end{figure}

The resulting response is shown in Fig.~\ref{fig:motor_trajectory}.
The vehicle follows the commanded horizontal path between the three
waypoints, with deviations occurring mainly during transitions between
successive reference points. The altitude response follows the commanded
changes while the horizontal maneuver is performed. The three-dimensional
trajectory further illustrates the simultaneous horizontal and vertical
motion of the vehicle. This test demonstrates the ability of the motor
pulsation controller to modify the direction of the phase-synchronized
control action as the horizontal reference changes.

\subsection{Flap Pulsation Control}

The flap pulsation controller was initially evaluated using the same
fixed horizontal and vertical position references as the motor-based
controller. The horizontal reference was set to
$p_{x,\mathrm{ref}}=0$~m and $p_{y,\mathrm{ref}}=1$~m, while the
altitude reference was maintained at $p_{z,\mathrm{ref}}=1$~m. In this
case, the horizontal control action was generated by phase-synchronized
flap deflections rather than differential motor thrust.

\begin{figure}[t]
    \centering
    \includegraphics[width=0.93\columnwidth]
    {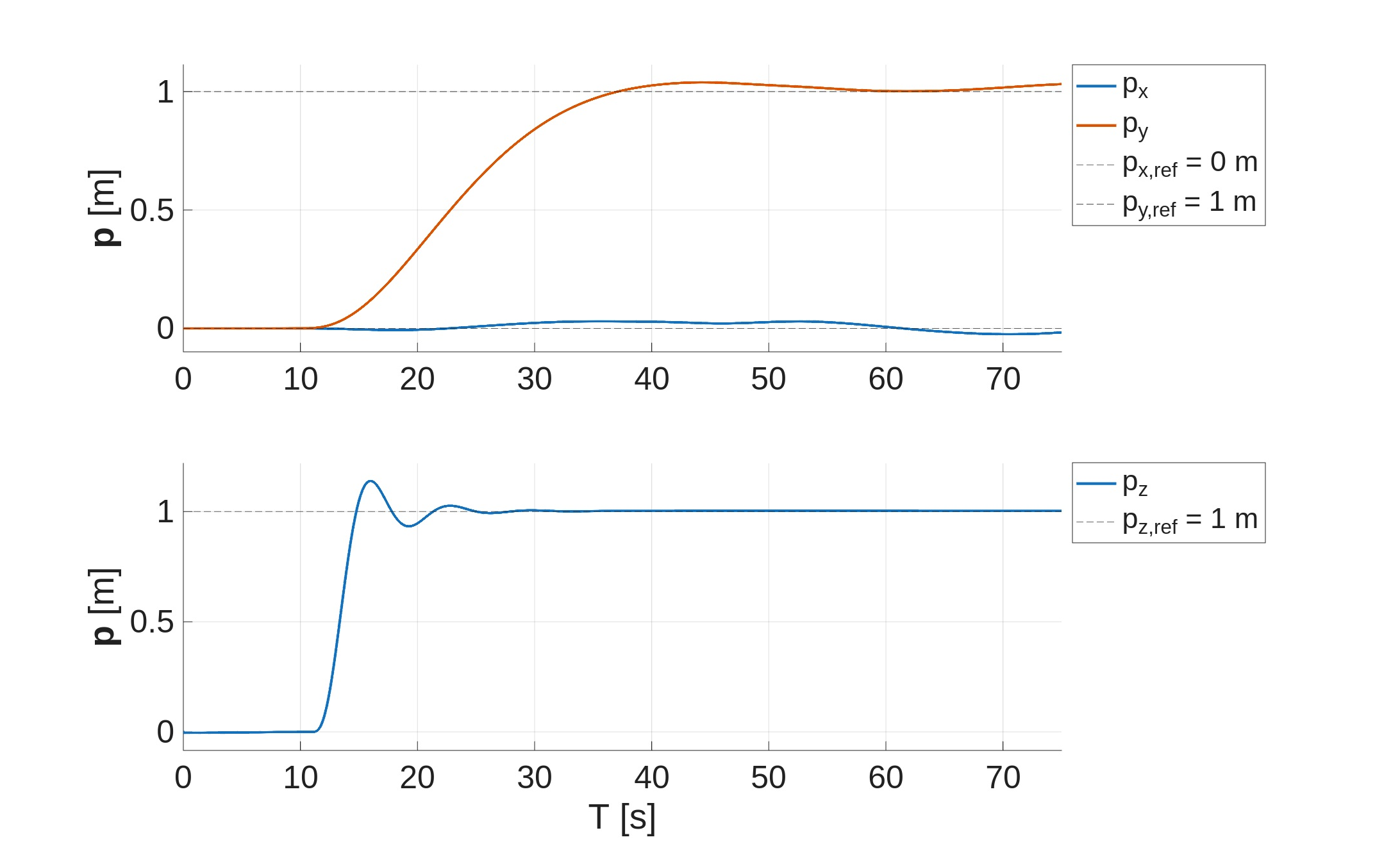}
    \caption{Position response using the phase-synchronized flap
    controller for $p_{x,\mathrm{ref}}=0$~m,
    $p_{y,\mathrm{ref}}=1$~m, and
    $p_{z,\mathrm{ref}}=1$~m.}
    \label{fig:flap_position}
\end{figure}

Fig.~\ref{fig:flap_position} shows the resulting position response.
The vehicle progresses toward the commanded $y$ position and approaches
the $1$~m reference, while the $x$ position remains close to its zero
reference. Following the initial transient, the altitude converges
toward the commanded value. The response therefore demonstrates that
phase-synchronized flap actuation can generate a directed horizontal
control action while the altitude-control loop operates independently.

The flap controller was subsequently evaluated using the same waypoint
sequence used for the motor pulsation controller. The horizontal
reference was varied between $P_1$, $P_2$, and $P_3$, while the altitude
reference was changed from $1$~m to $1.2$~m and subsequently returned
to $1$~m.

\begin{figure}[t]
    \centering
    \includegraphics[width=\columnwidth]
    {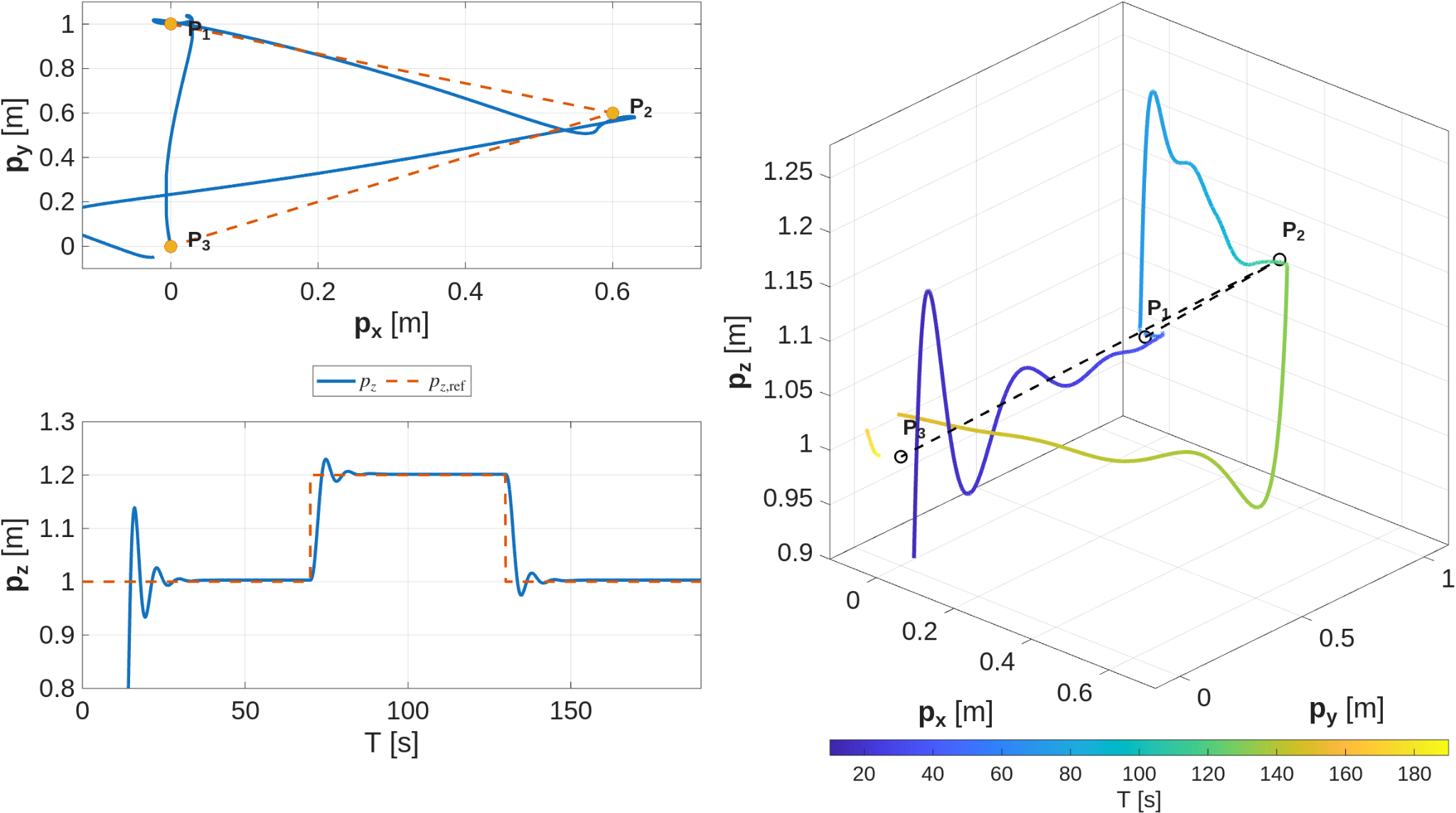}
    \caption{Waypoint tracking using the phase-synchronized flap
    controller. The horizontal trajectory is defined by waypoints
    $P_1$, $P_2$, and $P_3$, while the altitude reference is varied
    during the maneuver.}
    \label{fig:flap_trajectory}
\end{figure}

Fig.~\ref{fig:flap_trajectory} shows the resulting horizontal,
vertical, and three-dimensional responses. The vehicle follows the
commanded path between the successive waypoints, demonstrating that the
phase of the flap actuation can be adjusted as the desired horizontal
direction changes. Deviations from the commanded path are visible during
the transitions, while the vehicle approaches the vicinity of each
waypoint. The altitude response simultaneously follows the commanded
changes between \(1~\mathrm{m}\) and \(1.2~\mathrm{m}\).

The waypoint maneuver extends the fixed-reference test by requiring
changes in both the magnitude and direction of the horizontal control
action. The resulting trajectory demonstrates that phase-synchronized
aerodynamic actuation can provide horizontal position control for
successive reference directions while the motor system maintains the
required spinning-flight and altitude conditions.

The two control approaches achieve comparable horizontal waypoint-tracking performance, evaluated by the final waypoint error. Because the controllers are designed to reach successive reference positions rather than follow a prescribed trajectory, the evaluation uses the final waypoint error rather than path-based metrics. The mean final-waypoint
error, calculated over the commanded waypoints $P_1$, $P_2$, and $P_3$, was \(0.0397~\mathrm{m}\) for the motor and \(0.0394~\mathrm{m}\) for the flap pulsation controller. These results demonstrate that both phase-synchronized actuation methods can generate controlled horizontal motion toward successive reference positions.



\section{Experimental Evaluation}
\label{sec:experiment}

This section validates the simulation model of Section~\ref{sec:design} against the physical prototype by comparing steady-state response to a commanded throttle step change. A propeller's thrust depends not only on rotational speed but also on the advance ratio $J = V_\infty/(nD)$, where $V_\infty$ is the inflow velocity at the propeller disk, $n$ the rotational speed, and $D$ the propeller diameter; as $J$ increases, the thrust coefficient, and thus the thrust produced at a given rotational speed, decreases. The comparison is restricted to steady state because the expected mismatch, a $J$ dependent reduction in propeller thrust, depends on the tangential velocity the propeller experiences due to the vehicle's own spin, $V_{tan} = r_m\omega_x$ (Section~\ref{subsec:actuator_layout}), which reaches a fixed value once the measured spin rate $\omega_x$ settles.

\subsection{Experimental Setup}
\label{subsec:expsetup}

Each test began at a 30\% throttle baseline, matching the initial condition used in simulation, after which a step command was applied to a target between 60\% and 80\% in 5\% increments and held until $\omega_x$ reached steady state. This range was bounded by the measurement system (Optitrack) volume: below 60\% the vehicle did not lift off, and above 80\% it climbed out of the tracking volume before vertical velocity settled. Spin rate and vertical velocity were measured not only with an optical marker-tracking system, but were backed also by onboard IMU fusion. Three configurations were tested to separate wing angle and elevon effects: $30^\circ$ wing angle/$0^\circ$ flap (A), $40^\circ$ wing angle/$0^\circ$ flap (B), and $30^\circ$ wing angle/$15^\circ$ flap (C), each swept over the full throttle range described above. The UAV performing tests in a real-world scenario is shown in Fig. \ref{fig:metamorpher_real_world}.

\subsection{Results}
\label{subsec:results}

\begin{figure}[t]
\centering
\includegraphics[width=0.85\columnwidth]{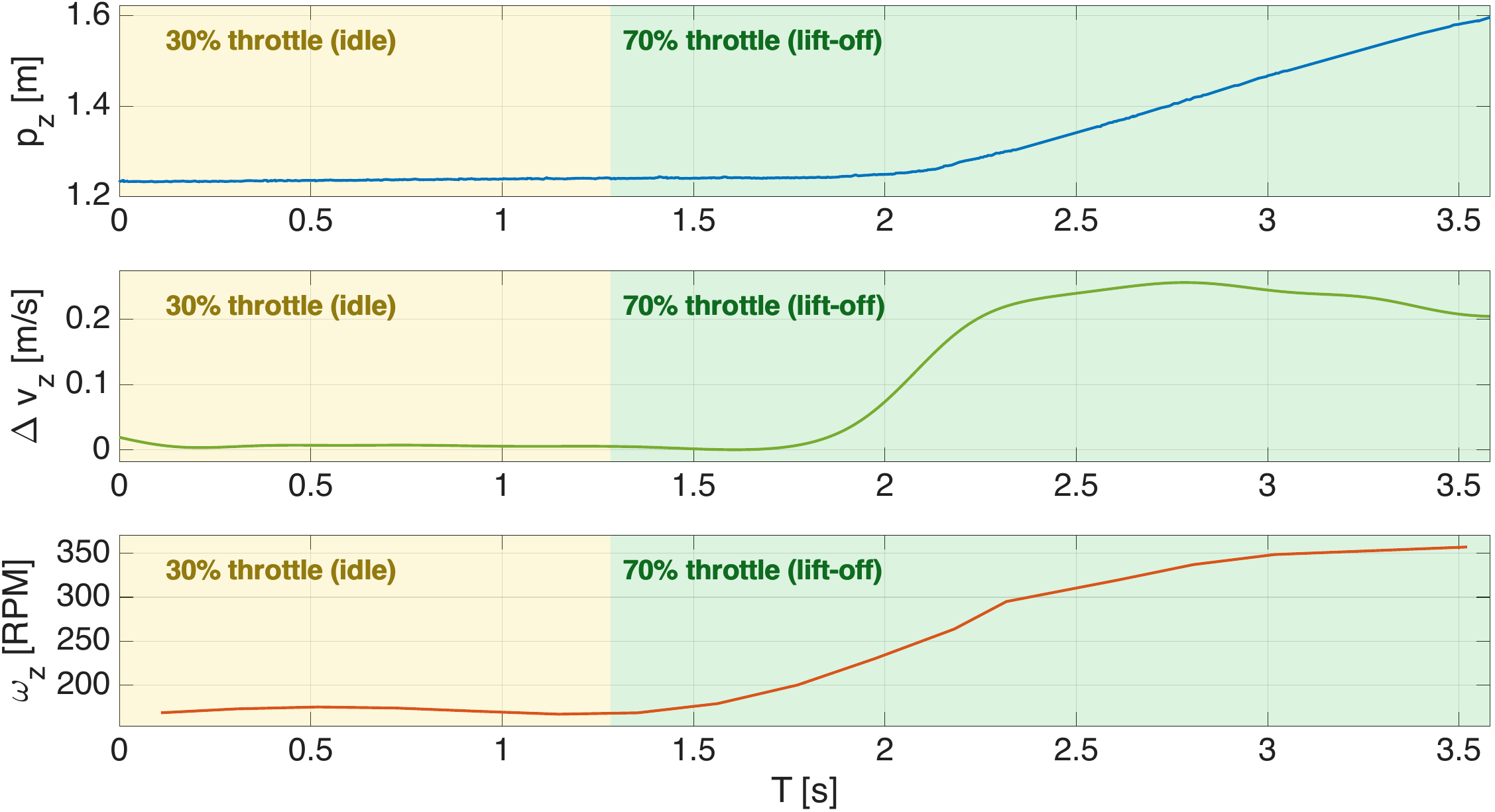}
\caption{Step response of configuration B showing vertical position, vertical velocity, and spin rate for a throttle step from 30\% to 70\%.}
\label{fig:stepresponse}
\end{figure}

Fig.~\ref{fig:stepresponse} shows a representative step from 30\% to 70\% throttle. Spin rate and vertical velocity both settle within roughly two seconds, with a mild overshoot in vertical velocity consistent with the second-order behaviour identified in simulation (Section~\ref{sec:simulation}); steady-state values used for analysis are taken from this settled portion of the response.

\begin{figure}[t]
\centering
\includegraphics[width=0.85\columnwidth]{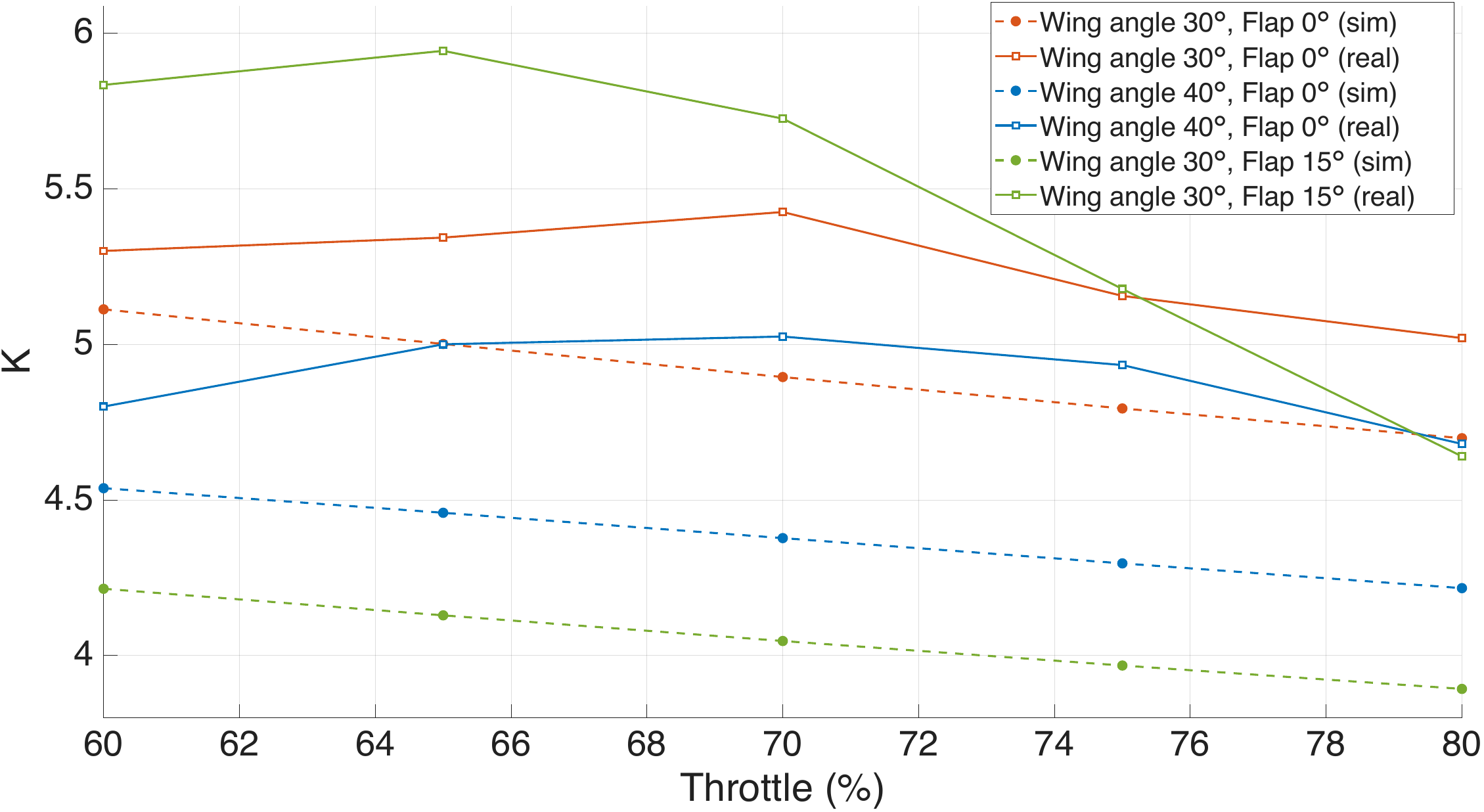}
\caption{Steady-state spin-rate gain versus throttle target for all three configurations, comparing simulation (dashed) against measured hardware response (markers).}
\label{fig:gainvsthrottle}
\end{figure}

To compare propulsion gain directly between the two domains, the simulation's throttle command is mapped linearly onto thrust, with 100\% throttle corresponding to $4$ N, the maximum static thrust the propeller/motor combination can produce at zero inflow velocity ($V_\infty = 0$). This common scaling places the simulated and measured steady-state spin-rate gain, $K = \Delta\omega_{ss}/\Delta u$, on the same axis in Fig.~\ref{fig:gainvsthrottle}. The simulated gain follows a smooth, monotonically decreasing trend. The measured gain instead departs from this trend and varies more between throttle levels, which is attributed to the advance-ratio-dependent reduction in thrust discussed above: the propeller does not deliver the full linearly-scaled thrust once it is spinning, an effect the simulation's static, linear assumption does not capture.

\begin{figure}[t]
\centering
\includegraphics[width=0.85\columnwidth]{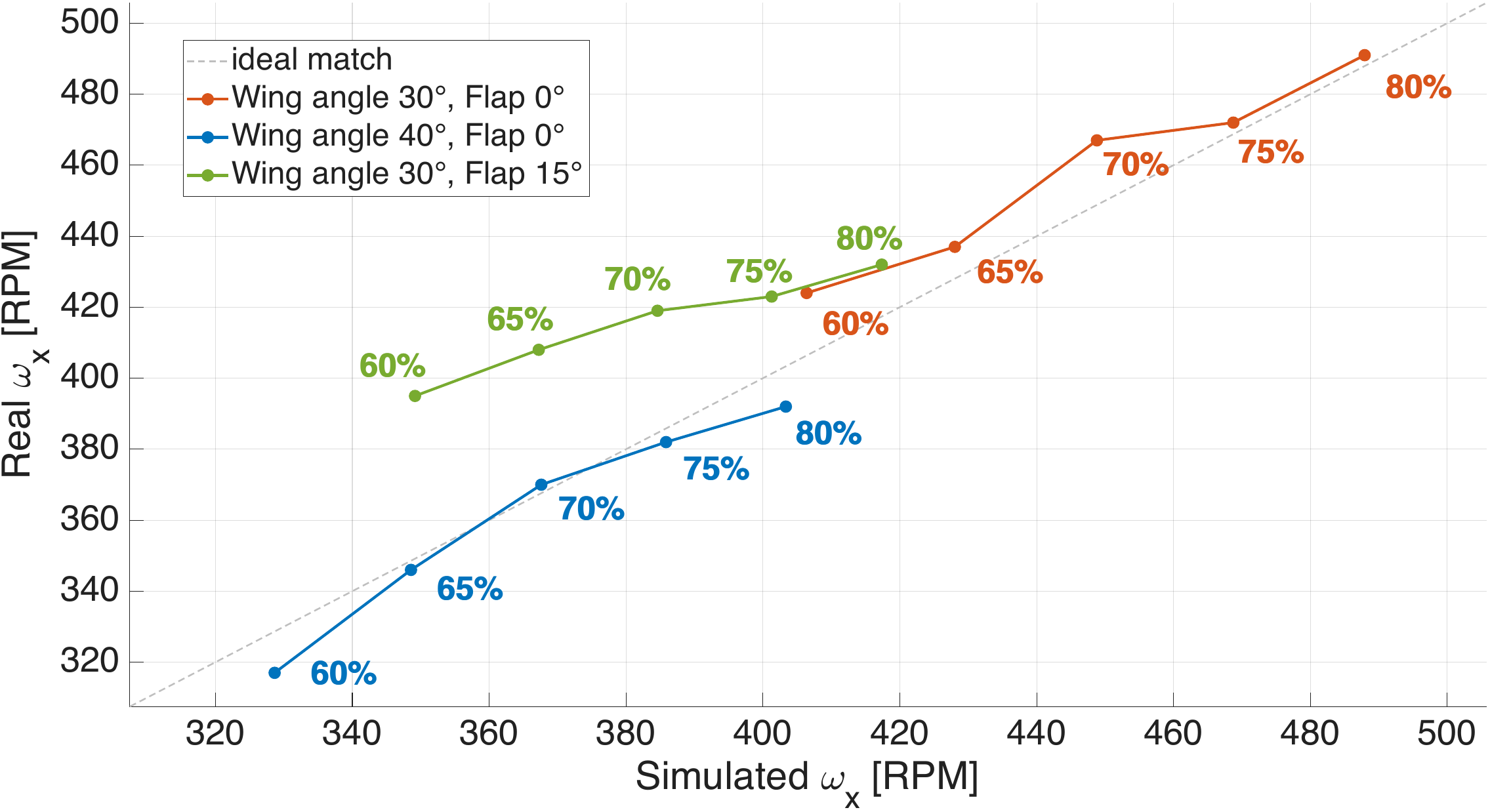}
\caption{Measured versus simulated steady-state spin rate for all three configurations, labeled by throttle target. The dashed line indicates ideal agreement.}
\label{fig:mapping}
\end{figure}

As a further consistency check on the propulsion model, Fig.~\ref{fig:mapping} compares measured against simulated steady-state spin rate for all three configurations (the perfect match is represented by the line). Configuration B matches simulation very closely (RMSE 2.1\%),
while configurations A and C, both at $30^\circ$, are 
slightly underpredicted by the model (RMSE 2.8\% and 9.1\%), with the gap slightly larger at low throttle and shrinking as throttle increases. This is consistent with the advance-ratio effect motivating the comparison, since the static thrust curve used in simulation does not account for the reduced thrust coefficient at nonzero inflow, the larger gap at $30^\circ$ than at $40^\circ$ supports this explanation without confirming it outright, since elevon-modified slipstream interaction could also contribute and is not isolated here.

\begin{figure}[t]
\centering
\includegraphics[width=0.85\columnwidth]{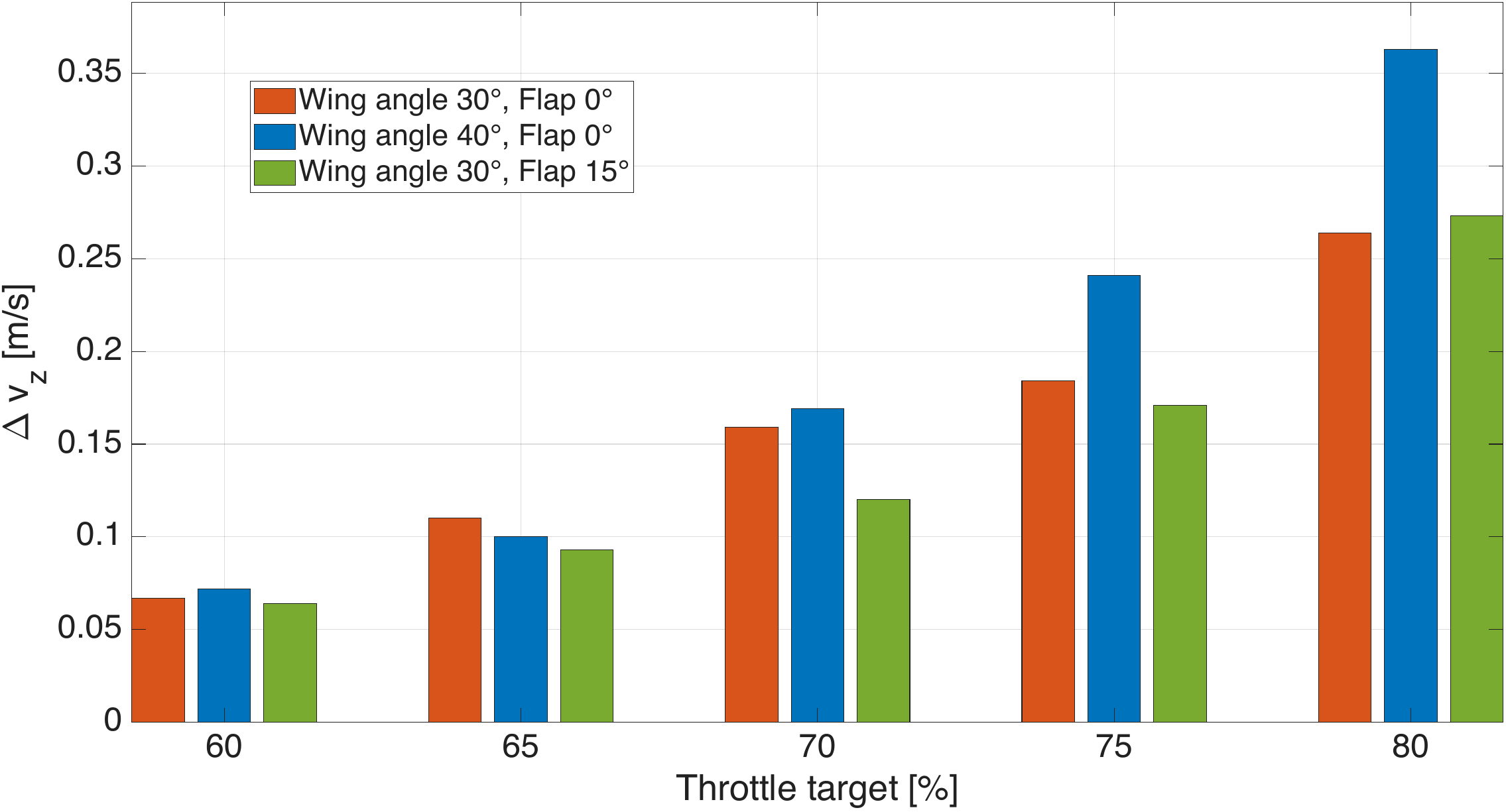}
\caption{Steady-state gain of vertical velocity with respect to throttle for all three configurations.}
\label{fig:zvelgain}
\end{figure}

Fig.~\ref{fig:zvelgain} compares the steady-state gain of vertical velocity to throttle across configurations. Unlike the spin-rate gain, this increases consistently with throttle for all three, and configuration B reaches the highest value at 80\% throttle, roughly 38\% above A and 33\% above C, indicating the $40^\circ$ wing angle converts throttle to climb rate most effectively. Because the propeller's tangential velocity remains an order of magnitude larger than the vehicle's vertical velocity even for the fastest-climbing configuration, the axial contribution to advance ratio stays small relative to the tangential component across all conditions tested, thus not impacting the propeller thrust.

This evaluation covers steady-state behaviour only. The phase-synchronized control strategies of Section~\ref{sec:control} act on timescales below the settling time characterized here, so validating the propulsion model's transient and periodic response remains a future work to be done before those strategies can be evaluated on hardware.
\section{Conclusion}
\label{sec:conclusion}


This paper presents the design and hover-mode control of MetaMorpher, a metamorphic UAV combining spinning-wing hover with fixed flying-wing cruise. Building on a previously validated flight dynamics model, we implemented a physical prototype with independently driven wing-mounted propellers and independently actuated elevons, and developed two phase-synchronized hover control strategies: motor pulsation, modulating differential thrust, and flap pulsation, a novel strategy pulsing elevon deflection while motor speed is held constant.
Simulation results validated both strategies, reaching comparable final-waypoint accuracy, suggesting the two channels are complementary rather than interchangeable. Experimental testing further validated the vertical-dynamics propulsion model against the physical prototype, showing very good steady-state agreement across three wing/elevon configurations.

Future work will address the propulsion model's transient and periodic response, a prerequisite for hardware evaluation of the horizontal motion controllers in hover mode. Also, synthesis of combined motor and flap pulsation into a single architecture exploiting their complementary strengths, should be investigated, which will provide a comparison of their efficiency and control performance across configurations by closed-loop flight testing of the full controller on the physical MetaMorpher prototype. The ultimate goal is to design and demonstrate the transition between spinning-wing hover mode and fixed-wing cruise mode to cover a complete flight envelope of MetaMorpher.

\appendices

\bibliographystyle{ieeetr}
\typeout{}
\balance
\bibliography{Bibliography/bibliography}

\end{document}